%% file: LOAD.tex
\pdfoutput=1

\documentclass[final,5pt,twocolumn]{elsarticle}

\def\etal{et al.}

\usepackage{times}
\usepackage{graphicx}
\usepackage{amsmath}
\usepackage{amssymb}
\usepackage{gensymb}
\usepackage{multirow}
\usepackage{amsmath,bm}
\usepackage[lofdepth,lotdepth]{subfig}

\usepackage{graphicx}
\usepackage{algorithm}
\usepackage{algpseudocode}

\newcommand{\sign}{\mathop{\mathrm{sign}}}

\usepackage{amssymb}

\begin{document}

\begin{frontmatter}



\title{LOAD: Local Orientation Adaptive Descriptor for Texture and Material Classification}


%
%
%
\author[1,2]{Xianbiao Qi}
\author[1]{Guoying Zhao}
\author[2]{Linlin Shen}
\author[2]{Qingquan Li}
\author[1]{Matti Pietik{\"a}inen}
%
\address[1]{Center for Machine Vision Research, University of Oulu, PO Box 4500, FIN-90014, Finland. E-mails: qixianbiao@gmail.com, gyzhao@ee.oulu.fi, mkp@ee.oulu.fi}
\address[2]{Shenzhen University, Shenzhen 518000, China. E-mails: llshen@szu.edu.cn, liqq@szu.edu.cn}
%
%
%



\begin{abstract}
{
   In this paper, we propose a novel local feature, called Local Orientation Adaptive Descriptor (LOAD), to capture regional texture in an image. In LOAD, we proposed to define point description on an Adaptive Coordinate System (ACS), adopt a binary sequence descriptor to capture relationships between one point and its neighbors and use multi-scale strategy to enhance the discriminative power of the descriptor. The proposed LOAD enjoys not only discriminative power to capture the texture information, but also has strong robustness to illumination variation and image rotation. Extensive experiments on benchmark data sets of texture classification and real-world material recognition show that the proposed LOAD yields the state-of-the-art performance. It is worth to mention that we achieve a 65.4\% classification accuracy-- which is, to the best of our knowledge, the highest record by far --on Flickr Material Database by using a single feature. Moreover, by combining LOAD with the feature extracted by Convolutional Neural Networks (CNN), we obtain significantly better performance than both the LOAD and CNN. This result confirms that the LOAD is complementary to the learning-based features.
}

\end{abstract}


\end{frontmatter}

\input{introduction.tex}

\input{relatedworks.tex}

\input{mainbody.tex}

\input{experiment.tex}
\input{conclusions.tex}
\bibliographystyle{elsarticle-harv}
\bibliography{egbib}


%
%
%
%

\end{document}

%% file: introduction.tex
\section{Introduction}
Visual image classification \cite{varma2003texture, varma2005statistical, liu2010exploring, sharan2013recognizing, foggia2013benchmarking, krizhevsky2012imagenet} is a challenging problem in computer vision, especially under multiple sources of image transformations, e.g. rotation, illumination, affine and scale variations, etc. The Bag-of-Words (BoW) \cite{csurka2004visual} model, as a powerful intermediate image representation of images, is the most popular approach in visual categorization in the past ten years.
In BoW model, the low-level feature extraction and mid-level feature encoding are two most important problems. In the past few years, some advanced middle-level feature encoding approaches has been proposed, such as Locality-constrained Linear Coding (LLC) \cite{wang2010locality}, Vector of Locally Aggregated Descriptors (VLAD) \cite{jegou2010aggregating} and Improved Fisher Vector (IFV) \cite{perronnin2010improving}.  These encoding methods have greatly put forward the development of BoW approach. However, on the other side, the development of low-level feature extraction is slow.

Earlier works on texture description mainly focused on capturing global texture information (e.g. GIST \cite{oliva2001modeling}, Gabor), or fine texture micro-structure (e.g. MR8 filter bank \cite{varma2005statistical}, Local Binary Pattern (LBP) \cite{ojala2002multiresolution}). The global texture descriptors can well capture global texture information, but miss most of texture details.
For instance, the GIST is good at capturing the spatial layout of scene, but performs poor on simple texture classification task in which the micro-structures are important.
These fine texture descriptors defined on very small patches (e.g. $3\times 3$ or $5\times 5$) can well capture small texture structures, but ignore global texture information. For example, the LBP and MR8 perform well on some simple texture data sets, but work poor on complex material data sets in which regional texture information is important.
There were some works that tried to bridge the gap between these two types of features. However, as we will discuss later, these features may suffer from some limitations, such as sensitiveness to image transformations or limited discriminative power.


This paper aims to provide a powerful regional texture descriptor. To this end, we propose a novel Local Orientation Adaptive Descriptor (LOAD). The proposed descriptor has two important advantages. (i) strong regional texture discrimination: the strong texture discrimination comes from two aspects. Firstly, on single point, we adopt a binary sequence description that owns stronger discriminative power than the Gradient Orientation in (e.g. SIFT \cite{lowe2004distinctive}, MORGH \cite{fan2012rotationally}) and Local Intensity Order (e.g. LIOP \cite{wang2011local}). Secondly, to enhance the discriminative power of the descriptor, we propose to use a multi-scale description to capture multi-scale texture information. (ii) robustness to image rotation and illumination variation: Due to that the LOAD is defined on an Adaptive Coordinate System, the LOAD is robust to image rotation. Meanwhile, the binary sequence description used in the LOAD affiliates the feature with great robustness to illumination variation.

Our {\bf{first contribution}} in this paper is to propose a novel and discriminative texture descriptor, LOAD, and demonstrate its effectiveness on two applications including texture and real-world material classification. On the traditional texture data sets \cite{ojala2002outex, lazebnik2005sparse}, the LOAD almost saturates the classification performance. On the real-world Flickr Material Database (FMD) \cite{liu2010exploring}, the LOAD achieves 65.4\% that is the best result for single feature as far as we know.

Our {\bf{second contribution}} is that we build a new real-world material data set from a newly introduced ETHZ Synthesizability data set. We name the newly introduced data set as OULU-ETHZ. We evaluate and compare the LOAD with the LBP, PRICoLBP and CNN on the OULU-ETHZ. Experiments show that our LOAD achieves promising performance on the new data set.

Our {\bf{third contribution}} is that we experimentally demonstrate that the proposed LOAD shows strong complementary property with the learning based feature, such as Convolutional Neural Networks (CNN) \cite{lecun1998gradient, krizhevsky2012imagenet}. On the Flickr Material Database \cite{liu2010exploring}, our LOAD combined with the CNN achieves 72.5\% that significantly outperforms the CNN (61.2\%) and LOAD (65.4\%). On the OULU-ETHZ data set, the combination of the LOAD and CNN improves the CNN by around 6.0\%.

We believe the strong complementary information is due to that the IFV representation with LOAD and CNN belong to two different approaches: non-structured and structured methods. The former is robust to image rotation and translation, but not well captures the structured information. In contrast, the latter is good at capturing the structured information because its hierarchical max-pooling strategy can preserve the structured information, but is not robust to heavy image rotation and translation.

%% file: relatedworks.tex
\section{Related Works}
Since the proposed descriptor is partially inspired by Local Binary Pattern (LBP) \cite{ojala2002multiresolution}, we will give a brief introduction to the LBP. 

\subsection{Local Binary Pattern}
\label{sec:LBP}


LBP is an effective gray-scale texture operator. Each LBP pattern corresponds to a kind of local structure in natural image, such as flat region, edge, contour and so on. For a pixel $(x_c, y_c)$ in an image $I$, its LBP image can be computed by thresholding the pixel values of its neighbors with the pixel value of the central point $(x_c, y_c)$:

\begin{equation}
LBP_{P, R}(x_c, y_c) = \sum_{p=0}^{P-1} \sign(g_p-g_c)2^p, \\
\sign{(t)} = \begin{cases} 1, & t \ge 0 \\
0, & t < 0, \\
\end{cases}
\label{eq:LBP}
\end{equation}
where $P$ is the number of neighbors and $R$ is the radius. $g_c = I(x_c, y_c)$ is the gray value of the central pixel $(x_c, y_c)$, and $g_p=I(x_p, y_p)$ is the value of its $p$-th neighbor $(x_p, y_p)$.
%

Ojala et al. \cite{ojala2002multiresolution} also pointed out that these patterns with at most two bitwise transitions described the fundamental properties of the image, and they called these patterns as ``uniform patterns''. The number of spatial transitions can be calculated as follows:
\begin{equation}
    \Phi{(LBP_{P, R}(x_c, y_c))} = \sum_{p=1}^{P} |\sign{(g_p-g_c)} - \sign{(g_{p-1}-g_c)}|,
\end{equation}
where $g_P$ equals to $g_0$. The uniform patterns are defined as $\Phi{(LBP(P, R))} \le 2$. For instance, ``11000011'' and ``00001110'' are two uniform patterns, while ``00100100'' and ``01001110'' are non-uniform patterns.

The LBP with $P = 8$ has $2^8=256$ patterns, in which there are 58 uniform patterns and 198 non-uniform patterns. According to the statistics in \cite{ojala2002multiresolution}, although the number of uniform patterns is significantly fewer than the non-uniform patterns, the ratio of uniform patterns accounts for 80\%-90\% of all patterns. Thus, instead of the original 256 LBP, the uniform LBP is widely used in many applications such as face recognition.

%% file: mainbody.tex
\begin{figure}
\begin{center}
\small
 \includegraphics[width=1.0\linewidth]{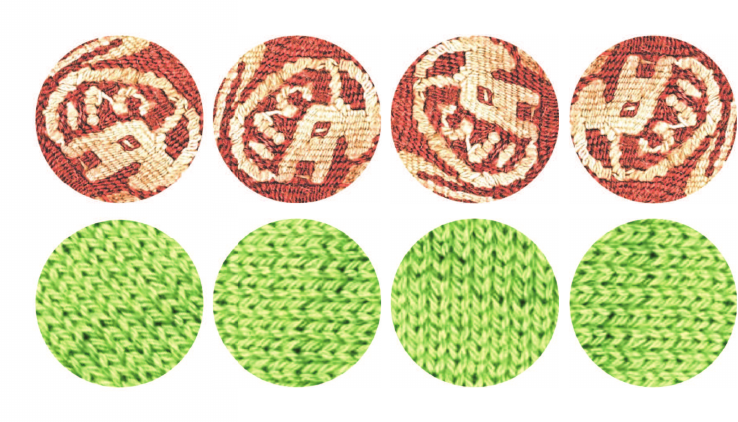}
\end{center}
   \caption{Sample patches under different image rotations.}
\label{fig:patches}
\end{figure}

\section{Local Orientation Adaptive Descriptor}
Our goal is to design a discriminative texture descriptor that owns the following two properties:
\begin{itemize}
\item {\bf{Regional texture discrimination: }} Most descriptors, such as SIFT, HOG$2\times 2$, are designed for image matching or human detection, not especially for texture description, thus their texture discrimination may be limited. Although there exist effective texture descriptors in literature, such as GIST, LBP, Completed LBP (CLBP), most of them are constructed for a global or fine texture description, thus they ignore regional texture information. In this work, we focus on designing a discriminatively regional texture descriptor.

\item {\bf{Robust to image transformations:}} Natural images contain rich image transformations, in which rotation and illumination variations are two most common cases. Thus, when designing a feature, these two aspects should be carefully considered.

\end{itemize}

In what follows we will describe the LOAD descriptor in detail. In Section \ref{sec:part1}, we describe the description strategy for each point under an adaptive coordinate system. Then in Section \ref{sec:part2}, we introduce a multi-scale description strategy that is used to enhance the discriminative power of the descriptor. And then, we describe the histogram construction and normalization approaches in Section \ref{sec:part3}. Finally, in Section \ref{sec:part4}, we discuss the relationship between the LOAD with some existing features.

\begin{figure}
\begin{center}
\small
 \includegraphics[width=0.95\linewidth]{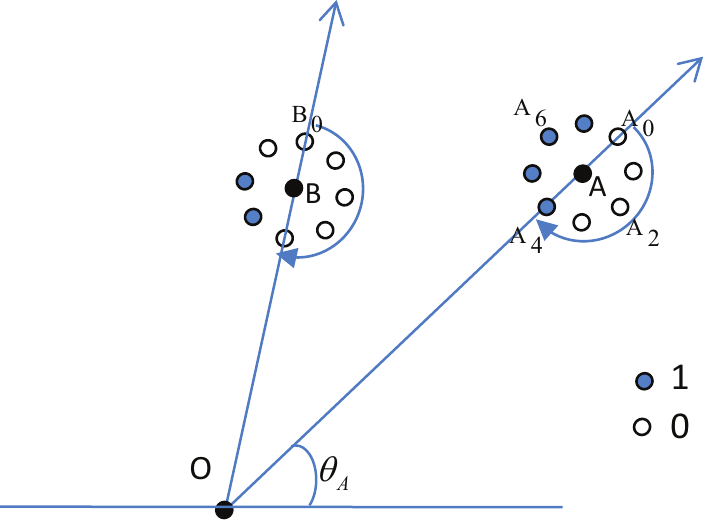}
\end{center}
   \caption{Illustration of Local Orientation Adaptive Descriptor. The point $O$ is the central point of the patch. The pattern for point $A$ is ``00001111'', and the pattern for point $B$ is ``00000110''.}
\label{fig:loadfig}
\end{figure}

\subsection{Point Description}
\label{sec:part1}
Given similar patches under different image rotations as shown in Figure \ref{fig:patches}, our objective is to extract a kind of descriptor that is discriminative and transformation invariant.
To achieve rotation invariance, the traditional methods (e.g. SIFT) firstly estimate a reference orientation (also called main orientation), and then align the patch to the reference orientation. However, estimation of the reference orientation will significantly increase the computational cost of the descriptors. Meanwhile, as indicated by \cite{fan2012rotationally}, the descriptor is sensitive to the error brought by the orientation estimation.

As the circular patch is symmetric with respect to any line across the central point, we choose to sample a circular region around each point. Given a sampled point $O$, we can obtain a circular patch around the point $O$.
By rotating the patch around the central point $O$, we can obtain a patch with arbitrary angle as shown in Figure \ref{fig:patches}. For any point $A$ in the patch, an Adaptive Coordinate System (ACS) can be formed by the point $A$ and the reference point $O$ as shown in Figure \ref{fig:loadfig}. Under the ACS, the neighboring relationship between point $A$ and its neighbors is invariant to image rotation. It means that, as shown in Figure \ref{fig:loadfig}, the positions of point $A$'s neighbors are always fixed compared to point $A$. Thus, the pixel values of the $A$'s neighbors are also invariant to image rotation.



Under the ACS, any point in the patch can be encoded in a rotation invariant way. In this paper, we propose a novel Local Orientation Adaptive Descriptor (LOAD) that is built on ACS. As illustrated in Figure \ref{fig:loadfig}, the LOAD pattern for the point $A$ can be encoded as follows:
\begin{equation}
\text{LOAD}_{P, R}(x_A, y_A, {\theta}_A) = \sum_{p=0}^{P-1} \sign{(V(A_p)-V(A))2^p}, \\\\\
\label{eq:LOADEq}
\end{equation}
\ \ \ \ \ \ \ \ \ \ \ \ \ \ \ $\begin{cases} x_{A_p} = x_A + Rcos(2\pi p/P - {\theta}_A), \\
y_{A_p} = y_A - Rsin(2\pi p/P - {\theta}_A), \\
\end{cases}$

where $P$ is the number of neighbors, $R$ is the radius, $(x_A, y_A)$ and $(x_{A_p}, y_{A_p})$ are the positions of the central point $A$ and its $p$-th neighbor under the ACS, $V(A)$ and $V(A_p)$ denote the pixel values of points $(x_A, y_A)$ and $(x_{A_p}, y_{A_p})$ individually, $\sign{(\cdot)}$ is a sign function, ${\theta}_A = \arctan \frac{y_A - y_O}{x_A - x_O}$.

In the same way, under the ACS, the adaptive gradient magnitude for the point $A$ can be denoted as follows:
\begin{equation}
\text{M}(A) = \sqrt[2]{(V(A_4)-V(A_0))^2 + (V(A_6)-V(A_2))^2}, \\\\\
\label{eq:magnitude}
\end{equation}
where the $\text{M}(A)$ is computed when $R = 1$.

The encoding approach as Eq. \ref{eq:LOADEq} has two advanced properties: (i) Rotation invariance: Under the ACS, the neighboring relationships between one point and its neighbors are fixed. As shown in Figure \ref{fig:loadfig}, the same start point $A_0$ will always be selected for the point $A$.
Thus, the LOAD encoding is rotation invariant. (ii) Robustness to illumination variation: Using the binary sequence description approach, our LOAD is also robust to illumination variation because illumination variation usually does not change the binary comparison relationship between two adjacent pixels.

According to Eq. \ref{eq:LOADEq}, when $P$ is set to 8, the LOAD will have 256 patterns that may be high for a local descriptor. Motivated by the ``Uniform'' encoding in LBP \cite{ojala2002multiresolution}, we also adopt the ``Uniform'' strategy the LOAD. Thus, the dimension of the LOAD is 59.

%
%
%
%
%
%
%

\begin{figure}
\begin{center}
\small
 \includegraphics[width=0.70\linewidth]{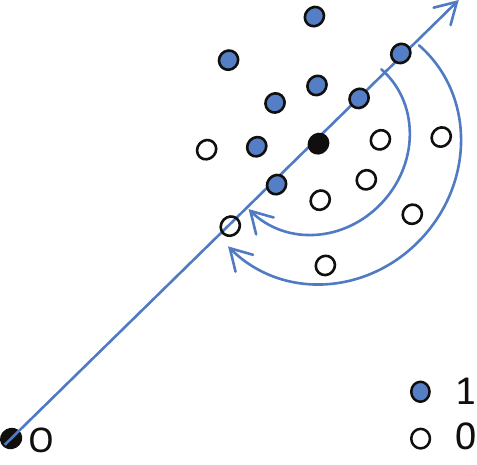}
\end{center}
   \caption{Multi-scale Local Orientation Adaptive Descriptor. The pattern for the inner scale is ``10001111'', and the pattern for the outer scale is ``10000011''.}
\label{fig:multiscaleM}
\end{figure}


\subsection{Multi-scale Description}
\label{sec:part2}

Multiresolution analysis--also called multi-scale analysis--is an effective way to depict texture information in different scales. Multi-scale strategy is widely used in the LBP \cite{ojala2002multiresolution} and its variants \cite{guo2010completed, guo2010rotation, zhao2012rotation}. As pointed out by previous works, the multi-scale description performs significantly better than the single-scale one. The multi-scale version of the LOAD can be defined as follows:
\begin{equation}
\text{LOAD}_{P, R}(x_A, y_A, {\theta}_A, s) = \sum_{p=0}^{P-1} \sign{(V(A_p)-V(A))2^p}, \\\\\
\label{eq:msLOAD}
\end{equation}
\ \ \ \ \ \ \ \ \ \ \ \ \ $\begin{cases} x_{A_p} = x_A + s \times Rcos(2\pi p/P - {\theta}_A), \\
y_{A_p} = y_A - s \times Rsin(2\pi p/P - {\theta}_A), \\
\end{cases}$

where $s$ is a scale factor.

Compared to the Eq. \ref{eq:LOADEq}, we introduce a scale factor to the Eq. \ref{eq:msLOAD}. With choice of different scale factors, we can obtain LOAD patterns in different scales. Figure \ref{fig:multiscaleM} shows the LOAD with two scales. In practice, we can choose 2, 3 or 4 scales. As shown in Figure \ref{fig:multiscaleM}, the binary sequence for the inner scale is ``10001111'', and the binary sequence for outer scale is ``10000011''. If the patterns between inner and outer scales are similar, it may indicate that the structures around this point is consistent, and vice versa.


\begin{algorithm}

  \caption{Calculation of LOAD feature}
  \begin{algorithmic}[1]

    \Require
      One reference point $O$ and a circular patch $P$ around $O$;
      
    \Ensure
      LOAD histogram feature $\text{H}$
    \State Initiate a 2-D histogram $\text{H}$ with zeros, the size of $\text{H}$ is set as $59\times S$;
    \For{all $O_i \in P$}
            \State Compute the gradient orientation $\text{M}(O_i)$ of the \\\ \ \ \ \ \ \ \ point $O_i$ as Eq. \ref{eq:magnitude},
            \For{each $s\in [1,S]$}
                \State Calculate the uniform LOAD pattern with $g_0$ \\\ \ \ \ \ \ \ \ \ \ \ \ \ \ \ as shown in Figure \ref{fig:loadfig} as start point, denote it \\\ \ \ \ \ \ \ \ \ \ \ \ \ \ \ as $\text{U}_{s}(O_i)$,
                \State Accumulate the histogram $\text{H}$, \\
                \ \ \ \ \ \ \ \ \ \ \ \ \ \ \ $\text{H}( \text{U}_{s}(O_i), s ) = \text{H}( \text{U}_{s}(O_i), s  ) + \text{M}(O_i)$,
            \EndFor
    \EndFor
    \State Resize the histogram $\text{H}$ into 1-D vector and normalize it with square root norm,
    \State Return $\text{H}$.
  \end{algorithmic}
  \label{alg:LOADApp}
\end{algorithm}

\subsection{Histogram Construction and Normalization}
\label{sec:part3}
Given a circular patch with the point $O$ as the central point, suppose that the patch has $K$ points. Assume that we use $S$ scales, the dimension for each scale is 59, thus, the final feature dimension is $59\times S$. We initiate a 2-D histogram $\text{H}$ with all zeros. Then, for each point $O_i,\ i \in [1, K]$, we can accumulate the histogram $\text{H}$ as follows:
\begin{equation}
\text{H}( \text{U}_{s}(O_i), s ) = \text{H}( \text{U}_{s}(O_i), s  ) + \text{M}(O_i), \\\\\
\label{eq:AULBP}
\end{equation}


where $s\in [1,S]$, $\text{M}(O_i)$ is the gradient magnitude of point $O_i$ under the ACS as computed according to the Eq. \ref{eq:magnitude}, $\text{U}_{s}(O_i)$ is the ``Uniform'' pattern of the LOAD feature of the point $O_i$ at the scale $s$.

After accumulating all $K$ points in the patch into the histogram $\text{H}$, we resize the histogram into 1-D vector.

Feature normalization is an important step for both feature description (e.g. RootSIFT \cite{arandjelovic2012three}) and image representation \cite{vedaldi2012efficient, perronnin2010improving}. In this paper, we follow the operator in RootSIFT, and conduct square root operation to our LOAD. Previous works \cite{arandjelovic2012three, vedaldi2012efficient} have shown that the square root normalization performs better than $L_2$ normalization.

For clarity, we summarize the algorithm for calculating the LOAD feature in Algorithm \ref{alg:LOADApp}, in which $S$ is the number of scales, $\text{U}_{s}(O_i)$ is the uniform pattern representation of the LOAD feature of the point $O_i$ at scale $s$.

\subsection{Relationship to Other Features}
\label{sec:part4}
Our LOAD feature is related to some existing features in the literature. The first category of related features are the LBP based methods, e.g. LBP \cite{ojala2002multiresolution}, CLBP \cite{guo2010completed}. Another set of related features are Local Intensity Order based methods including MORGH \cite{fan2012rotationally}, LIOP \cite{wang2011local}. However, different from the LBP based methods, our LOAD has the following two properties:
\begin{itemize}
\item Regional texture discrimination: Our LOAD is a patch-based feature. However, the LBP based methods, e.g. LBP, CLBP, were designed to depict micro-structures. Image representation based on LBP is to compute the histogram of patterns, but the image representation with the LOAD uses the BoW model.

\item Trade-off between rotation invariance and discriminative power: Our LOAD descriptor for each point is built on the ACS. Thus, the LOAD not only achieves good robustness to image rotation but also has strong discriminative power. On the other hand, the LBP based methods achieve rotation invariance at the cost of discriminative power.

\end{itemize}

Different from LIOP and MORGH, our LOAD has the following two properties:
\begin{itemize}
\item Richer patterns: The LOAD adopts a binary pattern description. Using the binary pattern descriptor, our LOAD has richer patterns than LIOP (16 patterns) and MORGH (8 patterns) on a single point.

\item Robust to the sensitiveness of region division: the LOAD does not employ the region division. Intensity order based region division \cite{fan2012rotationally, wang2011local} may be sensitive to non-monotonous illumination variation. Meanwhile, the region division will greatly increase the feature dimension.

\end{itemize}

%

\section{Encoding} The Improved Fisher Vector (IFV) \cite{perronnin2010improving} encoding has been proposed to address the problem of information loss in the process of feature encoding in the traditional BoW model. Within the context of IFV, images are represented by encoding densely sampled local descriptors. Principal Component Analysis (PCA) is firstly used to remove the correlation between two arbitrary dimensions. In PCA, we keep $D$ components. Then, a Gaussian mixture model (GMM) is estimated to build the visual words for the after-PCA local descriptors. The IFV measures the normalised deviations of local descriptors w.r.t. the GMM parameters.
More specifically, let $I =  \left \{\bm{x}_t, t = 1\cdots T  \right \}$ that are the set of \textit{$D$}-dimensional after-PCA local descriptors extracted from an image.
Denote the set of parameters of a \textit{K}-component GMM by $\lambda = \left \{ \pi_k, \mu_k, \Sigma_k, k=1,\cdots,K\right \}$, where $\pi_k$, $\mu_k$, and $ \Sigma_k$ are the prior, mean vector, and covariance matrix for the \textit{k}-th components respectively.
Given $\bm{x}_t$ with a soft assignment $\lambda_{tk}$ to each of the $K$ components, the IFV encoding of $I$ is defined as follows:
\begin{equation} \label{eq:fe_image}
\Phi \left ( I \right ) = \frac{1}{T}\sum_{t=1}^{T}\phi\left ( \bm{x}_t \right ),
\end{equation}
with
\begin{equation} \label{eq:fe_descriptor}
\phi \left ( \bm{x}_t \right ) = \left [ \phi_1\left ( \bm{x}_t \right ),\cdots,\phi_K\left ( \bm{x}_t \right )\right ],
\end{equation}
where
\begin{equation} \label{eq:fe_component}
\begin{split}
\phi_k\left ( \bm{x}_t \right ) = \left [ \underset{2D}{\underbrace{\frac{\lambda_{tk}}{\sqrt{\pi_k}}
\frac{\bm{x}_{t} - \mu_{k}}{\sigma_{k}},\frac{\lambda_{tk}}{\sqrt{2\pi_k}}
\left [\frac{(\bm{x}_{t} - \mu_{k})^2}{\sigma_{k}^2}-1\right ]} } \right ]^T, \\
 k = 1,\cdots,K.
\end{split}
\end{equation}
The IFV encoding is a vector representation of 2$D \times K$ dimensions. In the IFV, the power (signed square root) normalization usually shows better performance than the $\text{L}_2$ Normalization.

Compared with the BoW with K-means, the IFV framework provides a more general way to represent an image by a generative process of local descriptors and can be efficiently computed from much smaller vocabularies. Chatfield \etal \cite{chatfield2011devil} evaluated the state-of-the-art encoding methods such as the IFV, the Super Vector and the Locality-constrained linear (LLC), and showed that the IFV performs best in all compared encodings.

%% file: experiment.tex
\section{Experiments}
\subsection{Implementation Details}
{\bf{LOAD.}} In LOAD, we use four scales ((8, 1), (8, 2), (8, 3) and (8, 4)). The dimension for each scale is 59, thus, the final dimension is 236. Experiments show that the performance of four scales usually slightly improves the performance of two scales (e.g. (8, 1) and (8, 3)).

{\bf{IFV.}} We firstly sample 100,000 LOAD features from the training samples, then the
100,000 LOAD features are used to learn the PCA components, and 100 principal components are preserved as the basis for dimension reduction. As pointed out by \cite{sanchez2013image}, the PCA, which is used to remove correlation between two arbitrary dimensions, is a key step in the IFV framework. With the above-mentioned 100,000 after-PCA LOAD features, we learn a GMM with 256 components. For the PCA, we use the Matlab built-in SVD (Singular Value Decomposition). For the GMM, we use Vlfeat \cite{vedaldi2010vlfeat} to learn the parameters $\theta = \left \{ \pi_k, \mu_k, \Sigma_k, k=1,\cdots,K\right \}$. In the IFV, the $\Sigma_k$ is forced to be diagonal. The final dimension of the IFV representation for each image is $2\times100\times256 = 51,200$.

{\bf{Classifier.}} We trained a 1-vs-all linear SVM classifier (with C=10) using Liblinear \cite{fan2008liblinear} toolbox.

It should be pointed out that the computational cost for our LOAD descriptor is low. On a desktop computer with dual-core 3.4G CPU, the C++ (Matlab mex) implementation takes about 2s to extract 8000 features.

\subsection{Evaluation of Properties}

{\bf{Rotation Invariance.}} To evaluate the rotation invariance of the LOAD feature, we use three data sets: Outex\_TC\_00010 (TC10), Outex\_TC\_00012 (TC12) and UIUC. The experimental setups for each data set are presented in the following application section.
We compare the LOAD with RootSIFT. We guarantee that the LOAD and RootSIFT uses the same number of features and the same framework of IFV presentation. The experimental results for both features are shown in Tab. \ref{table:rotationEvaluation}.

\begin{table}[h]
\caption{Evaluation of Rotation Invariance of the LOAD on TC10, TC12 amd UIUC data sets.}       
\centering
\large
\begin{tabular}{ |c | c  |c | c |c |}
\hline
               &    UIUC    &    TC10   &  \multicolumn{2}{|c|}{TC12}   \\[0.02cm]
\hline
                    RootSIFT                       & 97.1  &  48.78     &  53.98 &  54.56      \\
 \hline
                    LOAD                           & 99.6  &  99.95     &  99.65 &  99.33      \\
 \hline
\end{tabular}
\label{table:rotationEvaluation}                   
\end{table}

According to Table \ref{table:rotationEvaluation}, we have two observations: (1). On the data sets with strong rotation such as TC10 and TC12, the LOAD shows great robustness to image rotation and significantly outperforms the RootSIFT. (2) On the UIUC data set that has small image rotations, our LOAD still shows better performance than the RootSIFT.

{\bf{Discriminative Power.}} To access the discriminative power of the LOAD, we directly compare it with the RootSIFT. We compare them in two sampling strategies: single-scale and multi-scale sampling. For single-scale, we directly sample points on the original images. For multi-scale sampling, we densely extracted features from six scales with rescaling factors $2^{-i/2}, i = -1, 0, 1, ..., 4$. We evaluate the LOAD and RootSIFT on Flickr Material Database (FMD) and UIUC data sets. The results are shown in Table \ref{table:discriminationEvaluation}.

\begin{table}[h]
\caption{Comparison of the LOAD and RootSIFT on FMD and UIUC data sets.}       
\centering                                     
\small
\begin{tabular}{ |c | c | c | c |}
\hline
     Sampling Strategy          &    Features    &    FMD     & UIUC  \\[0.02cm]
\hline
 \multirow{2}{1in}{Single-scale} & RootSIFT &  $56.5$        &  96.1     \\
                                 & LOAD &  $\bf{62.1}$        &  \bf{99.3}     \\
 \hline
 \multirow{2}{1in}{Multi-scale} & RootSIFT &  $60.5 $        &  97.1     \\
                                 & LOAD &  $\bf{65.4}$        &  \bf{99.6}     \\
\hline

\end{tabular}
\label{table:discriminationEvaluation}                   
\end{table}

From Table \ref{table:discriminationEvaluation}, on both single-scale and multi-scale sampling strategies, our LOAD outperforms the RootSIFT. For instance, with single-scale sampling, our LOAD improves the RootSIFT by 5.6\% on FMD data set. Meanwhile, we can also find that the multi-scale sampling strategy consistently outperforms the single-scale sampling strategy.

\begin{figure}
\begin{center}
\small
 \includegraphics[width=0.95\linewidth]{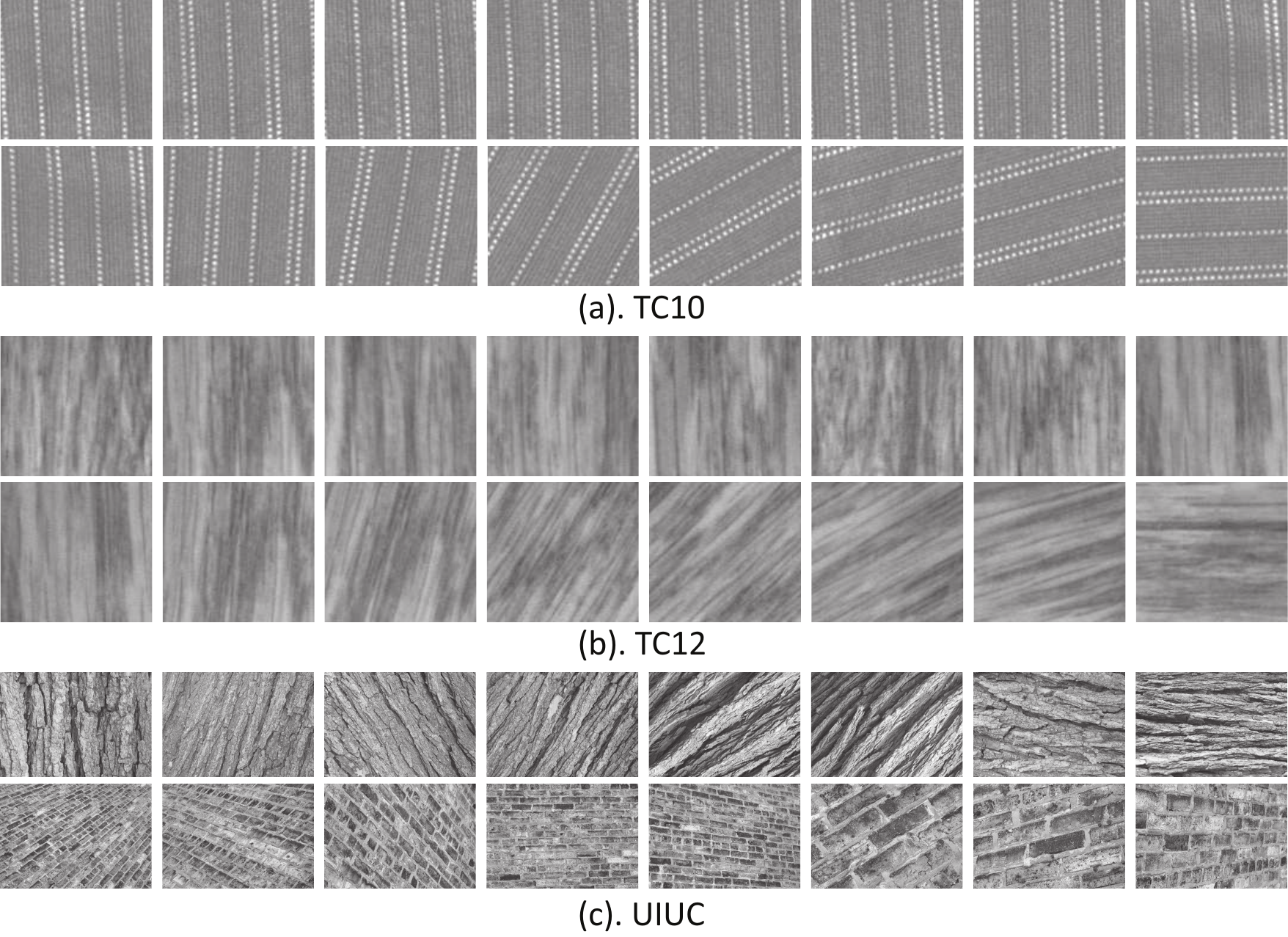}
\end{center}
   \caption{Sample images from TC10, TC12 and UIUC texture data sets. Note that TC10 and TC12 have strong rotation variation, and UIUC has strong rotation, scale and affine transformation.}
\label{fig:TC1012}
\end{figure}

\subsection{Texture Classification}
{\bf{Outex}} \cite{ojala2002outex} database has two test suites-{\bf{Outex\_TC\_00010 (TC10)}} and {\bf{Outex\_TC\_00012 (TC12)}}. The two test suites contain the
same 24 classes of textures, which were collected under three different illuminations (¡°horizon,¡± ¡°inca,¡± and ¡°t184¡±) and nine different rotation
angles (0, 5, 10, 15, 30, 45, 60, 75, and 90 ). There are 20 non-overlapping $128 \times 128$ texture samples for each class.
For TC10, samples of illuminations ``inca'' with angle 0 in each class were used for training and the other eight rotation angles
with the same illuminations were used for testing. Hence, there are 480 ($24 \times 20$) training samples and 3,840 ($24\times 20 \times 8$) validation samples.
For TC12, the classifier was trained with the same training samples as TC10, and it was tested with all samples captured under illuminations
 ``t184'' or ``horizon''. Hence, there are 480 ($24 \times 20$) training samples and 4,320 ($24\times 20 \times 9 $) validation samples for each illumination.
 It should be noted that the training images come from only one angle, but the testing images come from different angles.

{\bf{UIUC}} \cite{lazebnik2005sparse} texture data set contains 1,000 images: 25 different texture categories with 40 samples in each category. The image size in the data set is $640\times 480$. This data set has strong rotation and scale variations.
In the experiments, 20 samples from each category are used for training, and the rest 20 samples are used for testing.

Sample images for above three data sets are shown in Fig. \ref{fig:TC1012}. For all three data sets, we densely extracted features from six scales with rescaling factors $2^{-i/2}, i = -1, 0, 1, ..., 4$. We use the IFV representation and linear SVM. The results of TC10, TC12 and UIUC data sets are shown in Table \ref{tab:TC10TC12UIUC}.

%

\begin{table}[ht]
\centering
\normalsize
\subfloat[Subtable 1 list of tables text][Experimental results on data sets TC10 and TC12.]{
\begin{tabular}{ |c | c | c |c |}
\hline
Methods                 &    TC10   &  \multicolumn{2}{|c|}{TC12}   \\[0.02cm]
\hline
Dense SIFT (SVM)             &  48.78   &  53.98    &  54.56     \\[0.02cm]
\hline
CLBP\_SM/C (NN) \cite{guo2010completed}               &  99.14   &  95.18    &  95.55     \\[0.02cm]
\hline
BRINT (NN)  \cite{liu2013brint}                &  99.35   &  97.69    &  98.56     \\[0.02cm]
\hline
BRINT (SVM)  \cite{liu2013brint}                &  99.30   &  98.13    &  98.33     \\[0.02cm]
\hline
LOAD  (SVM)                  &  \bf{99.95}   &  \bf{99.65}    &  \bf{99.33}     \\[0.02cm]

\hline

\end{tabular}}
\qquad

\normalsize
\subfloat[Subtable 3 list of tables text][Experimental results on UIUC data set.]{
\begin{tabular}{|c | c | c | c |}
\hline
 Methods &  Acc. & Methods  & Acc. \\
\hline
   Lazebnik \etal \cite{lazebnik2005sparse}               &  96.0   & WMFS \cite{xu2010new}  & 98.6 \\
\hline
    BIF   \cite{crosier2010using}                 &  98.8   & SRP \cite{liu2011sorted}&  98.56 \\
\hline
  Sifre \etal \cite{sifre2013rotation}              & \bf{99.4}    &     RootSIFT     &  97.0        \\
\hline
  Cimpoi \etal \cite{Cimpoi14}   & 99.0    &     LOAD      &\bf{99.6}\\
 \hline
\end{tabular}}
\caption{Comparison with state-of-the-art methods on TC10, TC12 and UIUC texture data sets.}
\label{tab:TC10TC12UIUC}
\end{table}


Table \ref{tab:TC10TC12UIUC}(a) shows that the rotation invariant methods including CLBP, BRINT and LOAD significantly outperform the rotation sensitive method (Dense SIFT with IFV). Meanwhile, among all rotation invariant methods, our LOAD works best. According to Table \ref{tab:TC10TC12UIUC}(b), on UIUC data set, our LOAD also outperforms the state-of-the-art methods including SRP \cite{liu2011sorted} and two newly published works \cite{sifre2013rotation, Cimpoi14}.

%

\subsection{Real-World Material Classification}
\begin{figure}
\begin{center}
\small
 \includegraphics[width=0.95\linewidth]{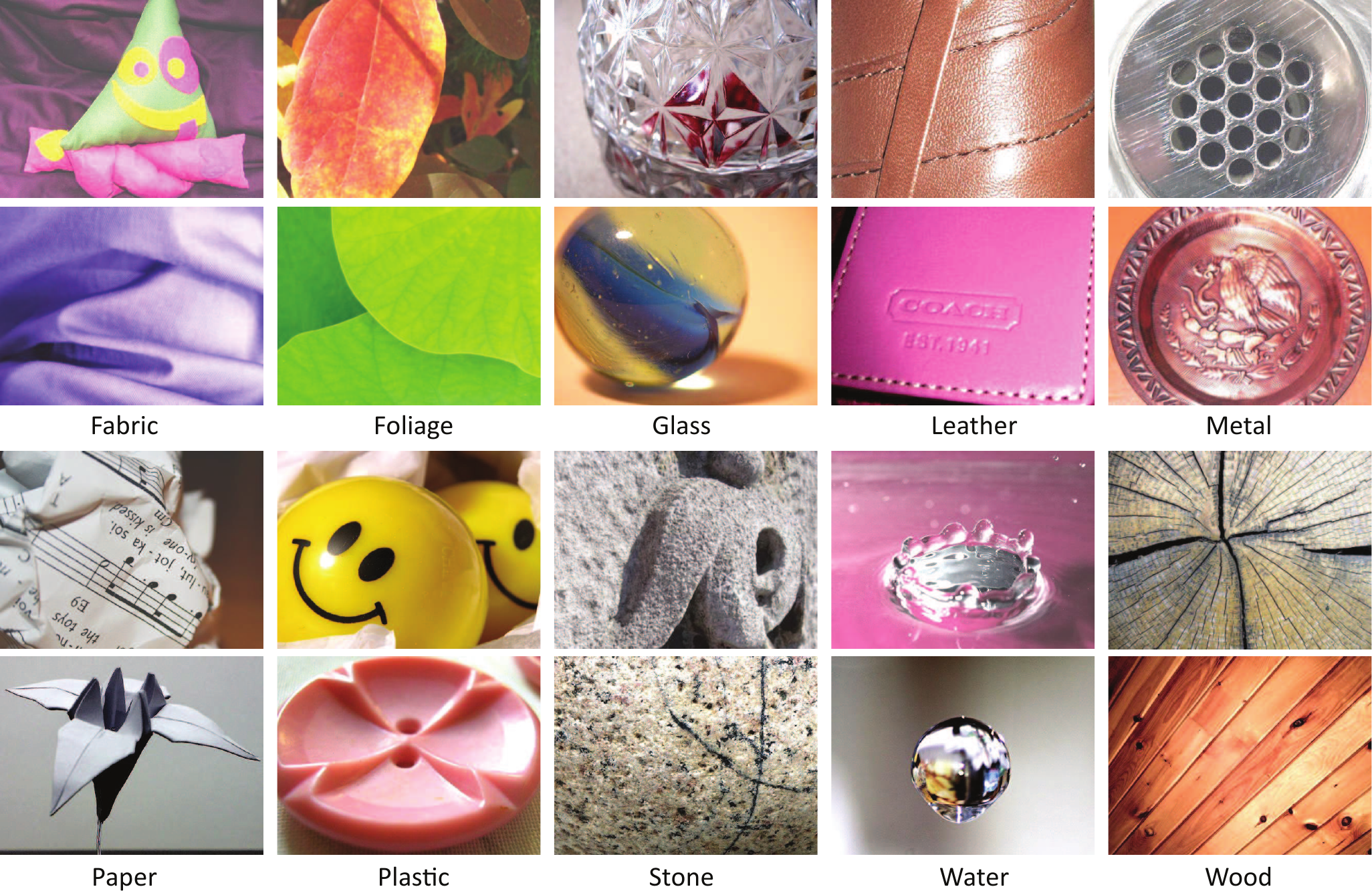}
\end{center}
   \caption{Sample images of 10 categories from the FMD data set.}
\label{fig:fmduiuc}
\end{figure}

{\bf{Flickr Material Dataset }}(FMD) \cite{liu2010exploring} is a challenging real-world material data set. It contains 10 categories, including fabric, foliage, glass, leather, metal, paper, plastic, stone, water, and wood. As pointed out in \cite{sharan2013recognizing}, FMD was designed with specific goal of capturing the appearance variations of real-world materials, and by including a diverse selection of samples in each category. Each category in FMD has 100 images, where 50 images are used for training and the rest 50 images are used for testing. Samples images are shown in Figure \ref{fig:fmduiuc}. We use the multi-scale sampling and densely extracted features from six scales with rescaling factors $2^{-i/2}, i = -1, 0, 1, ..., 4$. The step size for our sampling is 4. For instance, about 43,000 points are sampled from each image.


%
%

%
%

%

\begin{figure*}[t]
\centering
\includegraphics[width=1.0\linewidth]{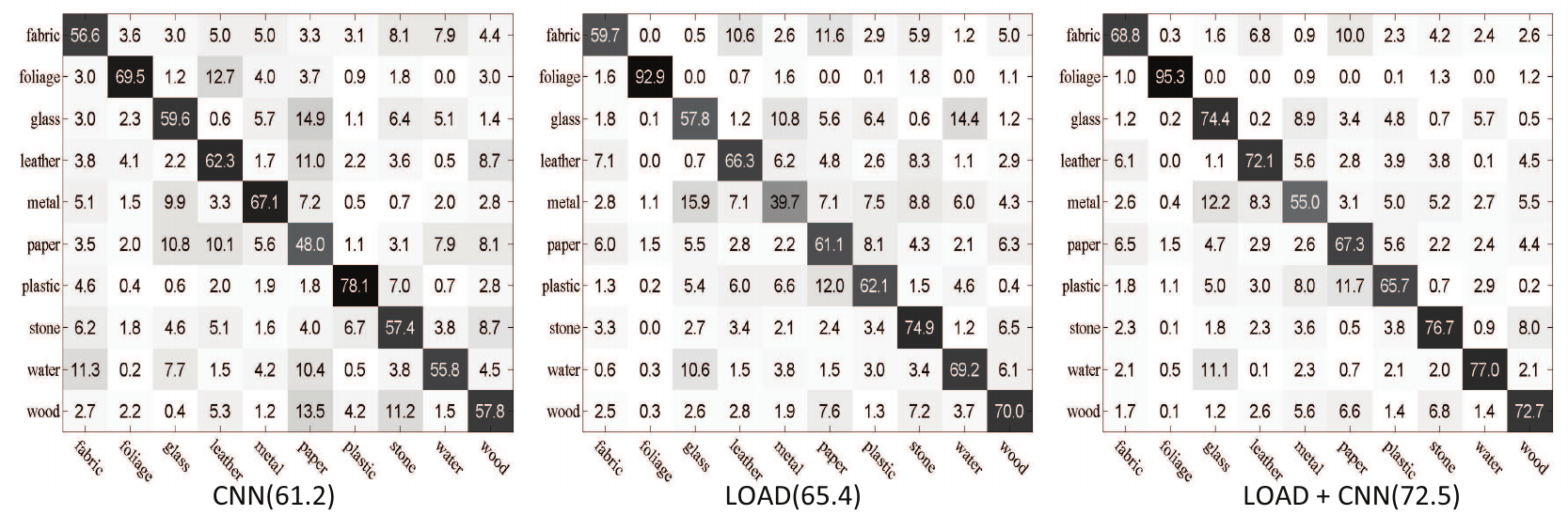}
\caption{Classification confusion matrices for CNN, LOAD and the combination of CNN and LOAD on FMD data set.}
\label{fig:fmdconfusionmatrices}
\end{figure*}

In the experiments, we compare our feature with many state-of-the-art methods including Kernel Descriptor \cite{hutoward}, Pairwise Rotation Invariance Co-occurrence LBP (PRICoLBP) \cite{qi2012pairwise}, DTD (a texture attribute descriptor) \cite{Cimpoi14}  and CNN\footnote{We use OverFeat\cite{sermanet2013overfeat} toolbox in this paper.} \cite{lecun1998gradient} and etc.

This paper investigates two key issues: (1) how much does the proposed feature depend on the dictionary (Learned by GMM) in IFV? (2) how much complementary information can the learning-based methods (e.g. CNN) provide for the LOAD feature with the IFV representation? For the first question, we compare the LOAD with the IFV representation using the vocabularies learned from the FMD or from an external data set. We randomly select 500 images from \cite{Dai14} as the external data set.
For the second question, we evaluate the combination of our LOAD with the CNN feature. All relevant results are shown in Table \ref{table:FMDState}, and three classification confusion matrices for the CNN, the LOAD and the combination of the CNN and the LOAD are shown in Figure \ref{fig:fmdconfusionmatrices}.

\begin{table}[h]
\caption{Comparison of state-of-the-art methods on FMD data set. LOAD* means using vocabulary learning from
an external data set. Note that the recognition accuracy for humans on the FMD is 84.9\% reported in \cite{sharan2013recognizing}.}       
\centering                                     

\begin{tabular}{ c  c  }
\hline
 Methods &  Accuracy \\
\hline
    Liu \etal CVPR'10 \cite{liu2010exploring}   &     44.6\\
 \hline
     Hu et al. BMVC'11 \cite{hutoward}    &    49\\
 \hline
    Qi \etal ECCV'12 \cite{qi2012pairwise}     &     $57.1\pm 1.8$ \\
 \hline
    Li \etal ECCV'12 \cite{li2012recognizing}    &     48.1\\
 \hline
    Sharan \etal IJCV'13 \cite{sharan2013recognizing} &     57.1\\
 \hline
   DTD CVPR'14 \cite{Cimpoi14} &     $49.8 \pm 1.3$ \\
 \hline
 Features Combined \cite{Cimpoi14} &     \bf{$67.1 \pm 1.5$}\\
 \hline
    CNN  \cite{sermanet2013overfeat}   &     $61.2 \pm 1.9$ \\
 \hline
        LOAD*        &     $64.6 \pm 1.7$ \\
 \hline
    LOAD        &     $\bf{65.4 \pm 1.7}$ \\
 \hline
    LOAD* + CNN        &     $\bf{72.5\pm 1.4}$ \\
 \hline
\end{tabular}

\label{table:FMDState}                   
\end{table}

From Table \ref{table:FMDState}, we can observe that:
\begin{itemize}
\item The LOAD achieves better performance than previous works including the methods with single feature, such as Kernel Descriptor, DTD, PRICoLBP. Meanwhile, it also outperforms some methods with multiple features, such as Liu \etal \cite{liu2010exploring} and Sharan \etal \cite{sharan2013recognizing}. Their results are based on combination of seven features.

\item The LOAD combined with the CNN significantly improves both of them. The combination of the CNN and LOAD decreases the error rate of LOAD by about 20\%, and decreases the error rate of CNN by about 30\%.

\item The LOAD is not sensitive to the source of the vocabulary. The LOAD with vocabulary learning from FMD only slightly improves the LOAD* with vocabulary learning from an external data set.

\end{itemize}

We can find that, from Figure \ref{fig:fmdconfusionmatrices}, the performances for the CNN and the LOAD on the corresponding categories vary a lot, such as the categories ``foliage'', ``metal'' and ``stone''. Meanwhile, we observe that on several categories, such as ``fabric'' and ``glass'', the LOAD combined with the CNN improves the one with lower classification accuracy by more than 10\%.

{\bf{Discussion.}} We believe the reason behind the significant increase of classification performance is that the CNN and IFV representations belong to two different approaches: structured and non-structured. The CNN is the structured method that is discriminative in capturing spatial layout information. With the hierarchical max-pooling strategy, the structured information is well preserved and captured.
However, on the other hand, the CNN may be not robust to heavy image rotation and translation. In contrast, the IFV representation with the LOAD feature is robust to image rotation and translation, but not powerful in describing spatial structure information. We believe this is the reason why these two methods have strong complementary information.


%
%
%
%
%
%
%
%
%


%

\begin{figure*}
\begin{center}
\small
 \includegraphics[width=0.98\linewidth]{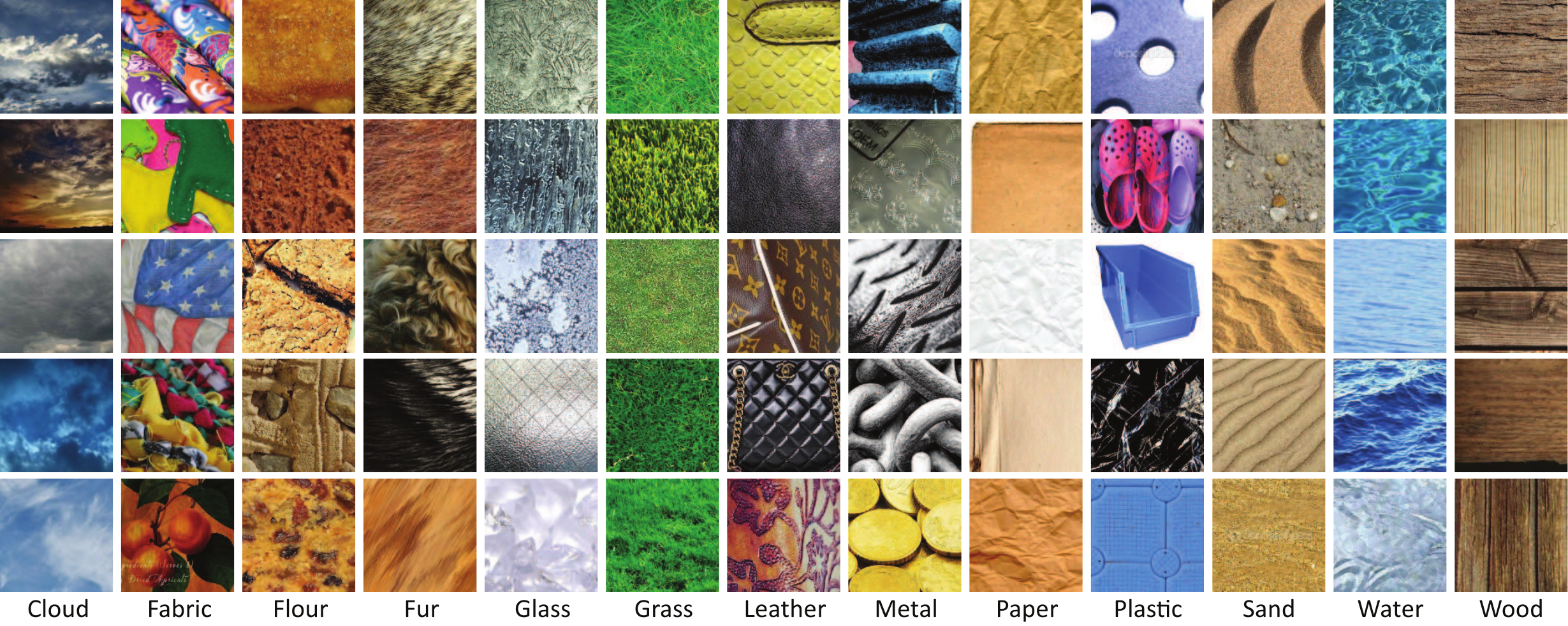}
\end{center}
   \caption{OULU-ETHZ real-world material data set. The OULU-ETHZ has rich image transformations.}
\label{fig:newdataset}
\end{figure*}

{\bf{A New Material Dataset (OULU-ETHZ) }} is introduced in this paper. The new data set is compiled from a new introduced ETHZ Synthesizability data set\footnote{The ETHZ Synthesizability data set contains 21302 texture of $300\times 300$ pixels, downloaded with 60 keywords.} that contains rich material images. The ETHZ data set is designed to evaluate the Synthesizability of images, but not designed for material recognition. In this paper, we select 13 material categories from this data set, and construct a new data set for material recognition.

All 13 categories include Cloud, Fabric, Flour, Fur, Glass, Grass, Leather, Metal, Paper, Plastic, Sand, Water and Wood. The number of the images in each category ranges from 44 to 420. Deriving from the ETHZ data set, the image sizes for all samples are $300\times 300$ pixels. Some sample images are shown in Figure
\ref{fig:newdataset}.

The OULU-ETHZ and FMD data sets share some similar properties and also have some differences. These similar and different properties are:
\begin{itemize}
\item The images in both FMD and OULU-ETHZ are both collected from real-world material images. Rich appearance variation happens in both data sets. For instance, the ``Air'' category in Figure \ref{fig:newdataset} has shown huge illumination variation.

\item Compared to the FMD data set, most of the images in the OULU-ETHZ are close-up images, thus, better alignment is shown in the OULU-ETHZ. It means that the images in the OULU-ETHZ has stronger scale and rotation prior than the FMD.

\end{itemize}


To evaluate different algorithms, we use 20 samples for training and the rest for testing. We pre-create five training-testing configurations, averaged accuracy is reported. We compare the proposed LOAD with two baseline methods (LBP, PRICoLBP) and also with CNN approaches\footnote{Following \cite{qi2012pairwise}, we use ${\chi}^2$ kernel for LBP and PRICoLBP. In the experiments, LBP uses three scales and PRICoLBP uses 6 templates. The dimensions for LBP and PRICoLBP are 54 and 3540 individually. We use linear SVM for our LOAD and CNN.}. The results are shown in Table \ref{table:ouluethz}.

\begin{table}[h]
\caption{Experimental results on the OULU-ETHZ set.}       
\centering                                     

\begin{tabular}{ c  c  }
\hline
 Methods &  Accuracy \\
 \hline
    LBP (Gray) \cite{ojala2002multiresolution}     &     $38.6\pm 1.2$ \\
 \hline
    PRICoLBP (Gray) \cite{qi2012pairwise}     &     $50.5\pm 1.8$ \\
 \hline
 PRICoLBP (Color) \cite{qi2012pairwise}     &     $52.8\pm 1.6$ \\
 \hline
    SIFT(IFV)        &     $53.2\pm 1.9$ \\
 \hline
    CNN        &     $62.1\pm 1.5$ \\
 \hline
    LOAD(IFV)        &     $55.9\pm 2.0$ \\
 \hline
    LOAD + CNN        &     $\bf{67.7\pm 1.6}$ \\
 \hline
\end{tabular}

\label{table:ouluethz}                   
\end{table}

From Table \ref{table:ouluethz}, we can observe that:
\begin{itemize}
\item The CNN achieves the best result among all compared approaches, our LOAD ranks second. The LOAD outperforms the LBP, PRICoLBP and SIFT. 

\item The LOAD shows strong complementary property with the CNN. The combination of them improves the CNN by about 6\%. 

\end{itemize}

{\bf{Discussion.}} It is interesting to investigate the reasons why the LOAD performs better than the CNN on the FMD, but worse than the CNN on OULU-ETHZ. We believe the following two points may be two main reasons:
\begin{itemize}

\item The OULU-ETHZ shows better consistency in appearance (e.g. color). The CNN is built on color image, and the LOAD is extracted from gray image. We believe that the OULU-ETHZ may have stronger color prior than the FMD. This argument can be validated by the fact that color PRICoLBP shows better performance than gray PRICoLBP on the OULU-ETHZ, but only achieves similar performance as gray PRICoLBP on the FMD. The consistency of appearance on the OULU-ETHZ is important for the CNN.

\item Most of the images in the OULU-ETHZ are close-up images. The close-up images have strong alignment on scale. Meanwhile, due to the skews when collecting the ETHZ data set, the images also have good alignment on rotation. The scale and rotation are two difficult issues to handle in the CNN.
\end{itemize}


%
%
%
%

%% file: conclusions.tex
\section{Conclusion}

This paper proposed a novel Local Orientation Adaptive Descriptor (LOAD) to capture regional texture information for image classification. It enjoys not only discriminative power to capture the texture information, but also has strong robustness to illumination variation and image rotation.
Superior performance on texture and real-world material classification tasks fully demonstrate its effectiveness. Meanwhile, it also shows strong complementary property with the learning-based method (e.g. Convolutional Neural Networks). The LOAD combined with the CNN significantly outperforms both of them. We believe the strong complementary information is due to that the IFV representation with LOAD feature and CNN belong to two different approaches: non-structured and structured approaches. The former is robust to image rotation and translation, but not well captures the structured information. In contrast, the latter is good at capturing the structured information because of its hierarchical max-pooling strategy, but is not robust to heavy image rotation and translation. Therefore, they exhibit strong complementary property.

%% file: LOAD.bbl
\begin{thebibliography}{38}
\expandafter\ifx\csname natexlab\endcsname\relax\def\natexlab#1{#1}\fi
\expandafter\ifx\csname url\endcsname\relax
  \def\url#1{\texttt{#1}}\fi
\expandafter\ifx\csname urlprefix\endcsname\relax\def\urlprefix{URL }\fi

\bibitem[{Arandjelovic and Zisserman(2012)}]{arandjelovic2012three}
Arandjelovic, R., Zisserman, A., 2012. Three things everyone should know to
  improve object retrieval. In: Computer Vision and Pattern Recognition.

\bibitem[{Chatfield et~al.(2011)Chatfield, Lempitsky, Vedaldi, and
  Zisserman}]{chatfield2011devil}
Chatfield, K., Lempitsky, V., Vedaldi, A., Zisserman, A., 2011. The devil is in
  the details: an evaluation of recent feature encoding methods. In: British
  Machine Vision Conference.

\bibitem[{Cimpoi et~al.(2014)Cimpoi, Maji, Kokkinos, Mohamed, and
  Vedaldi}]{Cimpoi14}
Cimpoi, M., Maji, S., Kokkinos, I., Mohamed, S., Vedaldi, A., 2014. Describing
  textures in the wild. In: Computer Vision and Pattern Recognition.

\bibitem[{Crosier and Griffin(2010)}]{crosier2010using}
Crosier, M., Griffin, L.~D., 2010. Using basic image features for texture
  classification. International Journal of Computer Vision.

\bibitem[{Csurka et~al.(2004)Csurka, Dance, Fan, Willamowski, and
  Bray}]{csurka2004visual}
Csurka, G., Dance, C., Fan, L., Willamowski, J., Bray, C., 2004. Visual
  categorization with bags of keypoints. In: Workshop on statistical learning
  in computer vision, European Conference on Computer Vision.

\bibitem[{Dai et~al.(2014)Dai, Riemenschneider, and Gool}]{Dai14}
Dai, D., Riemenschneider, H., Gool, L.~V., 2014. The synthesizability of
  texture examples. In: Computer Vision and Pattern Recognition.

\bibitem[{Fan et~al.(2012)Fan, Wu, and Hu}]{fan2012rotationally}
Fan, B., Wu, F., Hu, Z., 2012. Rotationally invariant descriptors using
  intensity order pooling. IEEE Transactions on Pattern Analysis and Machine
  Intelligence.

\bibitem[{Fan et~al.(2008)Fan, Chang, Hsieh, Wang, and Lin}]{fan2008liblinear}
Fan, R., Chang, K., Hsieh, C., Wang, X., Lin, C., 2008. Liblinear: A library
  for large linear classification. Journal of Machine Learning Research.

\bibitem[{Foggia et~al.(2013)Foggia, Percannella, Soda, and
  Vento}]{foggia2013benchmarking}
Foggia, P., Percannella, G., Soda, P., Vento, M., 2013. Benchmarking hep-2
  cells classification methods. IEEE Transaction on Medical Imaging.

\bibitem[{Guo et~al.(2010{\natexlab{a}})Guo, Zhang, and
  Zhang}]{guo2010completed}
Guo, Z., Zhang, L., Zhang, D., 2010{\natexlab{a}}. A completed modeling of
  local binary pattern operator for texture classification. IEEE Transactions
  on Image Processing.

\bibitem[{Guo et~al.(2010{\natexlab{b}})Guo, Zhang, and
  Zhang}]{guo2010rotation}
Guo, Z., Zhang, L., Zhang, D., 2010{\natexlab{b}}. Rotation invariant texture
  classification using lbp variance (lbpv) with global matching. Pattern
  Recognition.

\bibitem[{Hu and Bo(2011)}]{hutoward}
Hu, D., Bo, L., 2011. Toward robust material recognition for everyday objects.
  In: British Machine Vision Conference.

\bibitem[{J{\'e}gou et~al.(2010)J{\'e}gou, Douze, Schmid, and
  P{\'e}rez}]{jegou2010aggregating}
J{\'e}gou, H., Douze, M., Schmid, C., P{\'e}rez, P., 2010. Aggregating local
  descriptors into a compact image representation. In: Computer Vision and
  Pattern Recognition.

\bibitem[{Krizhevsky et~al.(2012)Krizhevsky, Sutskever, and
  Hinton}]{krizhevsky2012imagenet}
Krizhevsky, A., Sutskever, I., Hinton, G.~E., 2012. Imagenet classification
  with deep convolutional neural networks. In: Neural Information Processing
  Systems.

\bibitem[{Lazebnik et~al.(2005)Lazebnik, Schmid, and
  Ponce}]{lazebnik2005sparse}
Lazebnik, S., Schmid, C., Ponce, J., 2005. A sparse texture representation
  using local affine regions. IEEE Transactions on Pattern Analysis and Machine
  Intelligence.

\bibitem[{LeCun et~al.(1998)LeCun, Bottou, Bengio, and
  Haffner}]{lecun1998gradient}
LeCun, Y., Bottou, L., Bengio, Y., Haffner, P., 1998. Gradient-based learning
  applied to document recognition. Proceedings of the IEEE.

\bibitem[{Li and Fritz(2012)}]{li2012recognizing}
Li, W., Fritz, M., 2012. Recognizing materials from virtual examples. In:
  European Conference on Computer Vision.

\bibitem[{Liu et~al.(2010)Liu, Sharan, Adelson, and
  Rosenholtz}]{liu2010exploring}
Liu, C., Sharan, L., Adelson, E., Rosenholtz, R., 2010. Exploring features in a
  bayesian framework for material recognition. In: Computer Vision and Pattern
  Recognition.

\bibitem[{Liu et~al.(2011)Liu, Fieguth, Kuang, and Zha}]{liu2011sorted}
Liu, L., Fieguth, P., Kuang, G., Zha, H., 2011. Sorted random projections for
  robust texture classification. In: International Conference on Computer
  Vision.

\bibitem[{Liu et~al.(2014)Liu, Long, Fieguth, Lao, and Zhao}]{liu2013brint}
Liu, L., Long, Y., Fieguth, P., Lao, S., Zhao, G., 2014. Brint: Binary rotation
  invariant and noise tolerant texture classification. IEEE Transactions on
  Image Processing.

\bibitem[{Lowe(2004)}]{lowe2004distinctive}
Lowe, D., 2004. Distinctive image features from scale-invariant keypoints.
  International Journal of Computer Vision.

\bibitem[{Ojala et~al.(2002{\natexlab{a}})Ojala, Maenpaa, Pietikainen,
  Viertola, Kyllonen, and Huovinen}]{ojala2002outex}
Ojala, T., Maenpaa, T., Pietikainen, M., Viertola, J., Kyllonen, J., Huovinen,
  S., 2002{\natexlab{a}}. Outex-new framework for empirical evaluation of
  texture analysis algorithms. In: International Conference on Pattern
  Recognition.

\bibitem[{Ojala et~al.(2002{\natexlab{b}})Ojala, Pietik{\"a}inen, and
  M{\"a}enp{\"a}{\"a}}]{ojala2002multiresolution}
Ojala, T., Pietik{\"a}inen, M., M{\"a}enp{\"a}{\"a}, T., 2002{\natexlab{b}}.
  Multiresolution gray-scale and rotation invariant texture classification with
  local binary patterns. IEEE Transactions on Pattern Analysis and Machine
  Intelligence.

\bibitem[{Oliva and Torralba(2001)}]{oliva2001modeling}
Oliva, A., Torralba, A., 2001. Modeling the shape of the scene: A holistic
  representation of the spatial envelope. International Journal of Computer
  Vision.

\bibitem[{Perronnin et~al.(2010)Perronnin, S{\'a}nchez, and
  Mensink}]{perronnin2010improving}
Perronnin, F., S{\'a}nchez, J., Mensink, T., 2010. Improving the fisher kernel
  for large-scale image classification. In: European Conference on Computer
  Vision.

\bibitem[{Qi et~al.(2012)Qi, Xiao, Guo, and Zhang}]{qi2012pairwise}
Qi, X., Xiao, R., Guo, J., Zhang, L., 2012. Pairwise rotation invariant
  co-occurrence local binary pattern. European Conference on Computer Vision.

\bibitem[{S{\'a}nchez et~al.(2013)S{\'a}nchez, Perronnin, Mensink, and
  Verbeek}]{sanchez2013image}
S{\'a}nchez, J., Perronnin, F., Mensink, T., Verbeek, J., 2013. Image
  classification with the fisher vector: Theory and practice. International
  Journal of Computer Vision.

\bibitem[{Sermanet et~al.(2013)Sermanet, Eigen, Zhang, Mathieu, Fergus, and
  LeCun}]{sermanet2013overfeat}
Sermanet, P., Eigen, D., Zhang, X., Mathieu, M., Fergus, R., LeCun, Y., 2013.
  Overfeat: Integrated recognition, localization and detection using
  convolutional networks. arXiv preprint arXiv:1312.6229.

\bibitem[{Sharan et~al.(2013)Sharan, Liu, Rosenholtz, and
  Adelson}]{sharan2013recognizing}
Sharan, L., Liu, C., Rosenholtz, R., Adelson, E.~H., 2013. Recognizing
  materials using perceptually inspired features. International Journal of
  Computer Vision.

\bibitem[{Sifre and Mallat(2013)}]{sifre2013rotation}
Sifre, L., Mallat, S., 2013. Rotation, scaling and deformation invariant
  scattering for texture discrimination. In: Computer Vision and Pattern
  Recognition.

\bibitem[{Varma and Zisserman(2003)}]{varma2003texture}
Varma, M., Zisserman, A., 2003. Texture classification: Are filter banks
  necessary? In: Computer Vision and Pattern Recognition.

\bibitem[{Varma and Zisserman(2005)}]{varma2005statistical}
Varma, M., Zisserman, A., 2005. A statistical approach to texture
  classification from single images. International Journal of Computer Vision.

\bibitem[{Vedaldi and Fulkerson(2010)}]{vedaldi2010vlfeat}
Vedaldi, A., Fulkerson, B., 2010. Vlfeat: An open and portable library of
  computer vision algorithms. In: ACM Multimedia.

\bibitem[{Vedaldi and Zisserman(2012)}]{vedaldi2012efficient}
Vedaldi, A., Zisserman, A., 2012. Efficient additive kernels via explicit
  feature maps. IEEE Transactions on Pattern Analysis and Machine Intelligence.

\bibitem[{Wang et~al.(2010)Wang, Yang, Yu, Lv, Huang, and
  Gong}]{wang2010locality}
Wang, J., Yang, J., Yu, K., Lv, F., Huang, T., Gong, Y., 2010.
  Locality-constrained linear coding for image classification. In: Computer
  Vision and Pattern Recognition.

\bibitem[{Wang et~al.(2011)Wang, Fan, and Wu}]{wang2011local}
Wang, Z., Fan, B., Wu, F., 2011. Local intensity order pattern for feature
  description. In: International Conference on Computer Vision.

\bibitem[{Xu et~al.(2010)Xu, Yang, Ling, and Ji}]{xu2010new}
Xu, Y., Yang, X., Ling, H., Ji, H., 2010. A new texture descriptor using
  multifractal analysis in multi-orientation wavelet pyramid. In: Computer
  Vision and Pattern Recognition.

\bibitem[{Zhao et~al.(2012)Zhao, Ahonen, Matas, and
  Pietikainen}]{zhao2012rotation}
Zhao, G., Ahonen, T., Matas, J., Pietikainen, M., 2012. Rotation-invariant
  image and video description with local binary pattern features. IEEE
  Transactions on Image Processing.

\end{thebibliography}
